\pdfoutput=1
\documentclass[11pt]{article}

\usepackage{ACL2023}

\usepackage{times}
\usepackage{latexsym}

\usepackage[T1]{fontenc}
\usepackage[utf8]{inputenc}

\usepackage{microtype}

\usepackage{inconsolata}

\title{Zero-Shot Classification by Logical Reasoning on Natural Language Explanations}

\author{
Chi Han $^1$, Hengzhi Pei $^1$, Xinya Du $^2$, Heng Ji $^1$ \\
$^1$ University of Illinois at Urbana-Champaign \\
$^2$ The University of Texas at Dallas \\
\texttt{\{chihan3,hpei4,hengji\}@illinois.edu,~xinya.du@utdallas.edu}
}

\newcommand{\model}{CLORE}
\newcommand{\Model}{CLORE}

\usepackage{blindtext}
\usepackage{graphicx}
\usepackage{booktabs}
\usepackage{algorithm}
\usepackage{algorithmic}
\usepackage{color}
\usepackage{xcolor}
\usepackage{amsmath}
\usepackage{amsfonts}
\usepackage{makecell}
\usepackage[normalem]{ulem}
\usepackage{multirow}

\definecolor{OrangeRed}{rgb}{1.0, 0.27, 0.0}
\usepackage{xparse}

\NewDocumentCommand{\heng}{ mO{} }{}
\NewDocumentCommand{\chihan}{ mO{} }{}
\NewDocumentCommand{\hengzhi}{ mO{} }{}
\NewDocumentCommand{\zhenhailong}{ mO{} }{}
\NewDocumentCommand{\yi}{ mO{} }{}

\begin{document}
\maketitle
\begin{abstract}
    Humans can classify data of an unseen category by reasoning on its language explanations. This ability is owing to the compositional nature of language: we can combine previously seen attributes to describe the new category. For example, we might describe a sage thrasher
% \zhenhailong{ravens? (to be consistent with the intro)}
as "it has a slim straight relatively short bill, yellow
eyes and a long tail", so that others can use their knowledge of attributes ``slim straight relatively short bill'', ``yellow eyes'' and ``long tail'' to recognize a sage thrasher.
% \heng{\sout{you can consider revising it to be consistent with the example in Figure 1}}
Inspired by this observation, in this work we tackle zero-shot classification task by logically parsing and reasoning on natural language explanations. To this end, we propose the framework \model{} (Classification by LOgical Reasoning on Explanations). While previous methods usually regard textual information as implicit features, \model{} parses explanations into logical structures and then explicitly reasons along thess structures on the input to produce a classification score.
Experimental results on explanation-based zero-shot classification benchmarks demonstrate that \model{} is superior to baselines,
which we further show mainly comes from higher scores on tasks requiring more logical reasoning.
We also demonstrate that our framework can be extended to zero-shot classification on visual modality.
Alongside classification decisions, \model{} can provide the logical parsing and reasoning process as a clear form of rationale.
Through empirical analysis we demonstrate that \model{} is also less affected by linguistic biases than baselines.
% \heng{do you have results for this?} \hanchi{We do but the numbers are small, like 2\% for structured data, 3-8\% for text data and 1-4\% for visual data. Should we add them?}
% \yi{Need to finish converting template from EACL to ACL}
% \yi{\sout{is multimedia capability worthile to mention in abstract?}}
%     \footnote{Code and data will be made publicly available upon publication}
    \footnote{Code and data are available at \url{https://github.com/Glaciohound/CLORE}}
\end{abstract}

\section{Introduction}

\label{sec:intro}

\begin{figure}
\centering
\includegraphics[width=\columnwidth]{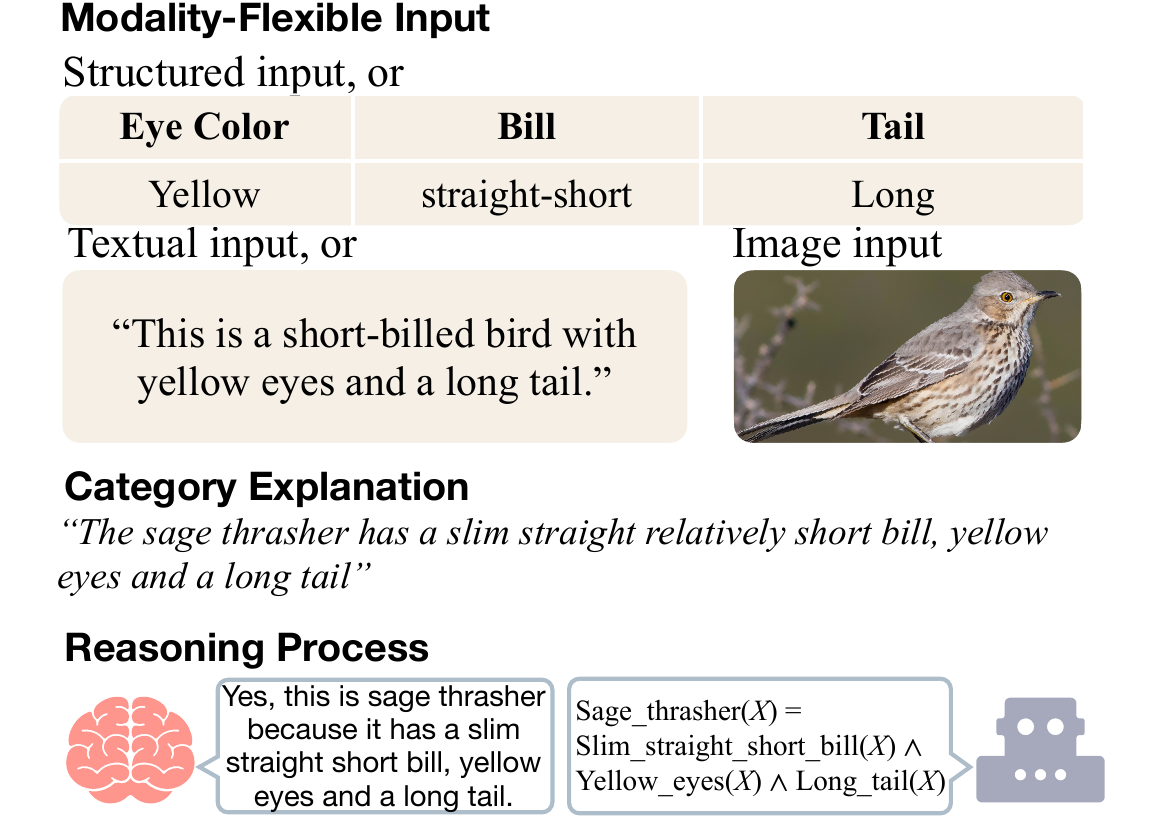}

\caption{We propose to conduct zero-shot classification by logical reasoning on natural language explanations, just like humans do. This design encourages our approach to better utilize the compositional property in natural language explanations.
% \yi{\sout{Would it make the figure any clearer by changing the top bold into something like "Input Probe", and the bottom bold as "Class Label Reference Explanation"}}
% \yi{\sout{Maybe add another bold section header of "Reasoning" right above the brain?}}
}

\vspace{-4mm}
\label{fig:teaser}
\end{figure}

\heng{maybe also show what SOTA caption generation methods can generate for the example in Figure 1.} \chihan{Emmmmm what is its role in the paper?}

\heng{\sout{In the text you are still talking about the old example. fix it}}
Humans are capable of understanding new categories by reasoning on natural language explanations~\citep{chopra2019first, tomasello2009cultural}.
For example, in Figure \ref{fig:teaser}, we can describe sage thrashers as ``having a slim straight relatively short bill, yellow eyes and a long tail''. Then when we view a real sage thrasher the first time, we can match its visual appearance with attributes ``slim straight relatively short bill'', ``yellow eyes'' and ``long tail'', and then logically combine these results to recognize it. % This requires learners to logically understand, parse and apply the explanations.
This ability has been shown to be applicable to both visual objects and abstract concepts~\citep{tomasello2009cultural}. Compared to learning only through examples, using language information enables humans to acquire higher accuracy in less learning time~\citep{chopra2019first}.
\heng{do you have results to show the performance on abstract concepts vs. concrete concepts?}\chihan{I can claim that the ``CLUES'' experiment is about abstract concepts, and ``CUB bird classification'' involves concrete concepts. (?) But I do not have a good idea on how to compare these two...}

\begin{figure*}[t]
\centering
\includegraphics[width=\textwidth]{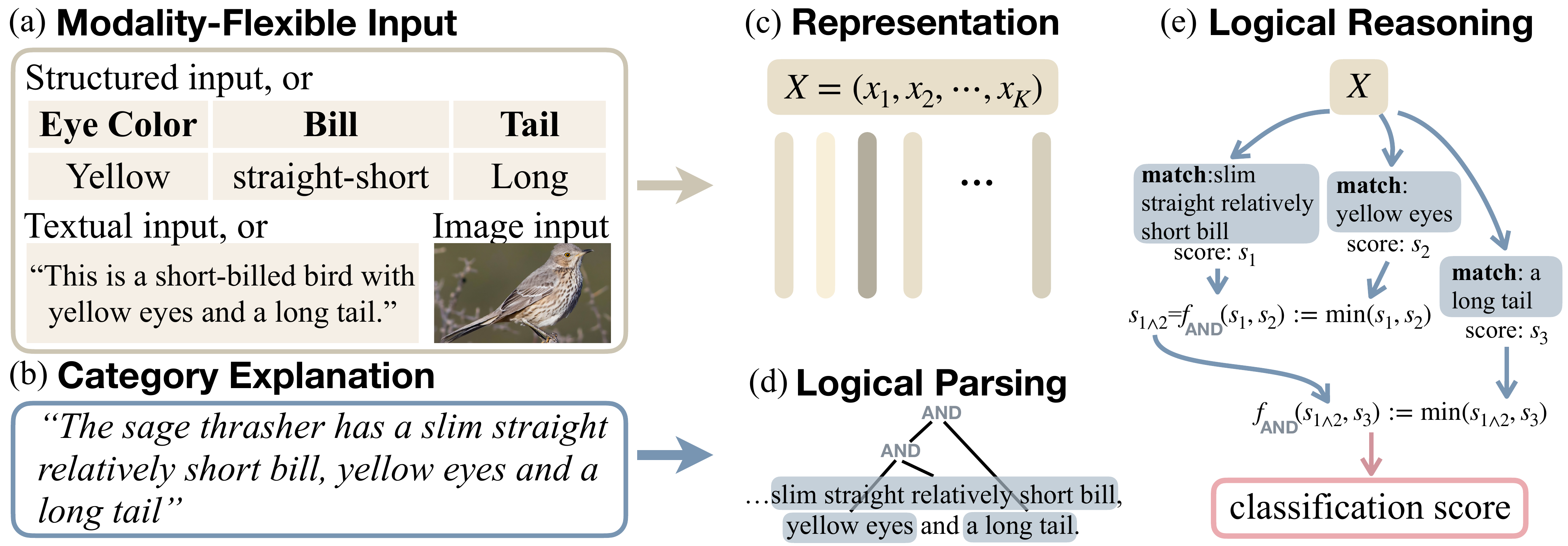}

\caption{An illustrative figure of \model{}'s working paradigm. After encoding the input (sub-figure(c)) we conduct logical parsing (sub-figure (d)) and logical reasoning (sub-figure(e)) over the explanations to obtain the classification score.
% \xd{you didn't ref Figure 2 until sec approach. maybe move it to page 3 if fig 1 can cover intro.}
% \hanchi{Heng: explain how what information is shared among explanations (the alignment between attributes and input features)}
}

\vspace{-4mm}
\label{fig:approach}
\end{figure*}

One important advantage of learning with natural language explanations is that explanations are often logical and compositional. That is, we can logically decompose the explanation of a new category into previously seen attributes (or similar ones) such as ``yellow eyes'' and ``long tail''. This enables us to reuse the knowledge on how these attributes align with visual appearances, and reduce the need for ``trial-and-error''.
Furthermore, learning with explanations provides better interpretability which makes results more trustworthy. 
% This property is of vital importance in transparency-sensitive domains, such as legal judgement, where some sort of ``rationales'' are preferred by human observers~\citep{branting2021scalable}.

Recently, there have been research efforts on using language information for zero-shot generalization. Types of such language information include human-annotated explanations or task-level instructions \citep{menon2022clues, sanh2022multitask, mishra2022cross}. However, auxiliary language information is often treated merely as additional text sequences to be fed into pre-trained language models. This approach does not fully leverage the compositional nature of natural language, and does not provide sufficient interpretable rationales for its decisions.
% \zhenhailong{\sout{This paragraph reads a little disconnected with the zero-shot classification task and our method; maybe change the order of this paragraph with the next paragraph: first talk about the task we are interested in, then introduce previous method's limitation/inspiration (this paragraph), then introduce our method?}}
% \chihan{I have trouble making a decision. @Heng can you give an advice on the ordering of this paragraph? Should we say ``previous methods were bad, so we do xxx'' or ``we do xxx, while previous methods were bad''?}

Inspired by these observations, in this work we explore classifying unseen categories by logically reasoning on their language explanations. To this end, we propose the framework of Classification by LOgical Reasoning on Explanations (\model{}). 
\model{} works in two stages: it first parses an explanation into a logical structure, and then reasons along this logical structure. Figure~\ref{fig:approach} illustrates an example of classifying sage thrashers in this way. We first encode the inputs (Figure~\ref{fig:approach} (a) $\rightarrow$ (c)) get the logical structure of explanation (Figure~\ref{fig:approach} (b) $\rightarrow$ (d)). Then we detect if the input matches attributes, and we gather the matching scores along the logical structure to output the overall classification score (Figure~\ref{fig:approach} (c),(d)$\rightarrow$(e)). In this case the logical structure consists of AND operators over three attributes.
We test the model's zero-shot capacity by letting it learn on a subset of categories, and make it categorize data from other unseen types.

We conduct a thorough set of analysis on the latest benchmark for zero-shot classifier learning with explanations, CLUES~\citep{menon2022clues}. %\zhenhailong{add citation?} 
% Our analysis shows that \model{} works better than baselines on tasks requiring higher level of compositional reasoning, which validates our model design. \model{} also demonstrates better interpretability and robustness against linguistic biases. Furthermore, as a test on generalizability of the proposed approach on other modalities, we built two new benchmarks on zero-shot classification with explanations: CUB-Explanations and ECtHR-Explanations. They are built upon the image dataset CUB-200-2011~\citep{wah2011caltech} and legal-text dataset ECtHR~\citep{chalkidis2021paragraph}, and we associate each category with a set of language explanations. \model{} consistently outperforms baseline models in zero-shot classification across modalities.
Our analysis shows that \model{} works better than baselines on tasks requiring higher level of compositional reasoning, which validates the importance of logical reasoning in \model{}. \model{} also demonstrates better interpretability and robustness against linguistic biases. Furthermore, as a test on generalizability of the proposed approach on other modalities, we built a new benchmark on visual domain: CUB-Explanations. It is built upon the image dataset CUB-200-2011~\citep{wah2011caltech}, while we associate each category with a set of language explanations. \model{} consistently outperforms baseline models in zero-shot classification across modalities.

% The rest of the paper is organized as follows:
% Section~\ref{sec:related} will list most related research areas to our work. Then we will describe our proposed approach in Section~\ref{sec:approach}. Finally in Section~\ref{sec:experiments} we will explain experiment settings and give result analysis.
To sum up, our contributions are as follows:
\begin{itemize}
    \item We propose a novel zero-shot classification framework by logically parsing and reasoning over explanations.
    % Empirical results demonstrate its superior performance than baseline models, especially on tasks that require more compositional reasoning.
    
    \item We demonstrate our model's superior performance and explainability, and empirically show that \model{} is more robust to linguistic biases and reasoning complexity than black-box baselines.
    
    % \item We demonstrate the universality of the proposed approach by building two new benchmarks, CUB-Explanations and ECtHR-Explanations. They are derived from CUB-200-2011~\citep{wah2011caltech} and ECtHR~\citep{chalkidis2021paragraph} by collecting natural language explanations for each category.

    \item We demonstrate the universality of the proposed approach by building a new benchmarks, CUB-Explanations. It is derived from CUB-200-2011~\citep{wah2011caltech} by collecting natural language explanations for each category.
    
\end{itemize}

\section{Related Work}

\label{sec:related}

% \heng{the '~' means a blank space, so if you use '~' you get two white blanks}
\paragraph{Classification with Auxiliary Information}

This work studies the problem of classification through explanations, which is related to classification with auxiliary information. For example, in the natural language processing field,~\citet{JMLR:v11:mann10a, ganchev2010posterior} incorporate side information (such as class distribution and linguistic structures) as a regularization for semi-supervised learning. Some other efforts convert crowd-sourced explanations into pseudo-data generators for data augmentation when training data is limited~\citep{wang2020learning, hancock2018training, DBLP:conf/iclr/WangQZ0YN0R20}. However, these explanations are limited to describing linguistic patterns (e.g., ``this is class X because word A directly precedes B''), and are only used for generating pseudo labels. A probably more related topic is using explanations for generating a vector of features for classification~\cite{srivastava2017joint, srivastava2018zero}. However, they either learn a black-box final classifier on features or rely on observed attributes of data, so their ability of generalization is limited. 

% \heng{\sout{Check Manling's CVPR22 paper and its related work, there are many related papers about using NL explanations for cross-media tasks}}
The computer vision area widely uses class-level auxiliary information such as textual metadata, class taxonomy and expert-annotated feature vectors~\citep{yang2022comprehensive, akata2015evaluation, xian2016latent, lampert2009learning, akata2015label, samplawski2020zero}. However, the use of label names and class explanations is mainly limited to a simple text encoder~\citep{akata2015evaluation, xian2016latent, liu2021goal, norouzi2014zero}. This processing treats every text as one simple vector in similarity space or probability space, whereas our method aims to reason on the explanation and exploit its compositional nature.
% Recently, there has been impressive progress on vision-language pre-trained models (VLPMs)~\citep{li2022clip, radford2021learning, li2019visualbert, kim2021vilt}. These methods are trained on large-scale high-quality vision-text pairs with contrastive learning~\citep{radford2021learning, kim2021vilt, li2019visualbert} or mask prediction objective~\citep{kim2021vilt, li2019visualbert}. 
% However, these model mostly focus on representation learning than understanding the compositionality in language.
% As we will show through experiments, VLPMs fits data better at the cost of zero-shot generalization performance.
% As we will show through experiments, this approach performs less well when the inputs go beyond simple pattern matching and require more reasoning. On the contrary, 

\paragraph{Few-shot and Zero-shot Learning with Language Guidance}
% \heng{\sout{There are a lot of work on zero-shot for NLP tasks, you are missing a lot, such as Lifu's early work: https://blender.cs.illinois.edu/paper/zeroshot2018.pdf or you can focus on cross-media zero-shot tasks only}}
This work deals with the problem of learning with limited data with the help of natural language information, which is closely related to few-shot and zero-shot learning with language guidance in NLP domain~\citep{hancock2018training, DBLP:conf/iclr/WangQZ0YN0R20, srivastava2017joint, srivastava2018zero, yu2022building, huang2018zero}. Besides the discussions in the previous subsection, recent pre-trained language models (LMs)~\citep{kenton2019bert, liu2019roberta, tam-etal-2021-improving, gao-etal-2021-making, yu2022building} have made huge progress in few-shot and zero-shot learning. To adapt LMs to downstream tasks, common practices are to formulate them as cloze questions~\citep{tam-etal-2021-improving, schick2021s, menon2022clues, li2022piled} or use text prompts~\citep{mishra2022cross, ye2021crossfit, sanh2022multitask, aghajanyan-etal-2021-muppet}. These approaches hypothetically utilize the language models' implicit reasoning ability~\citep{menon2022clues}. However, in this work we demonstrate with empirical evidence that adopting an explicit logical reasoning approach can provide better interpretability and robustness to linguistic biases.

In computer vision, recently there has been impressive progress on vision-language pre-trained models (VLPMs)~\citep{li2022clip, radford2021learning, li2019visualbert, kim2021vilt}. These methods are trained on large-scale high-quality vision-text pairs with contrastive learning~\citep{radford2021learning, kim2021vilt, li2019visualbert} or mask prediction objective~\citep{kim2021vilt, li2019visualbert}. 
However, these model mostly focus on representation learning than understanding the compositionality in language.
As we will show through experiments, VLPMs fits data better at the cost of zero-shot generalization performance.

There are also efforts in building benchmarks for cross-task generalization with natural language explanations or instructions~\citep{mishra2022cross, menon2022clues}. We use the CLUES benchmark~\citep{menon2022clues} in our experiment for structured data classification, but leave~\citet{mishra2022cross} for future work as its instructions are focused on generally describing the task instead of defining categories/labels.

\paragraph{Neuro-Symbolic Reasoning for Question Answering} is also closely related to our approach. Recent work~\citep{mao2019neuro, yi2018neural, han2019visual} has demonstrated its efficacy in question answering, concept learning and image retrieval. Different from our work, previous efforts mainly focus on question answering tasks, which contains abundant supervision for parsing natural language questions. In classification tasks, however, the number of available explanations is much more limited (100$\sim$1000), which poses a higher challenge on the generalization of reasoning ability.

\begin{figure}
\centering
\includegraphics[width=\columnwidth]{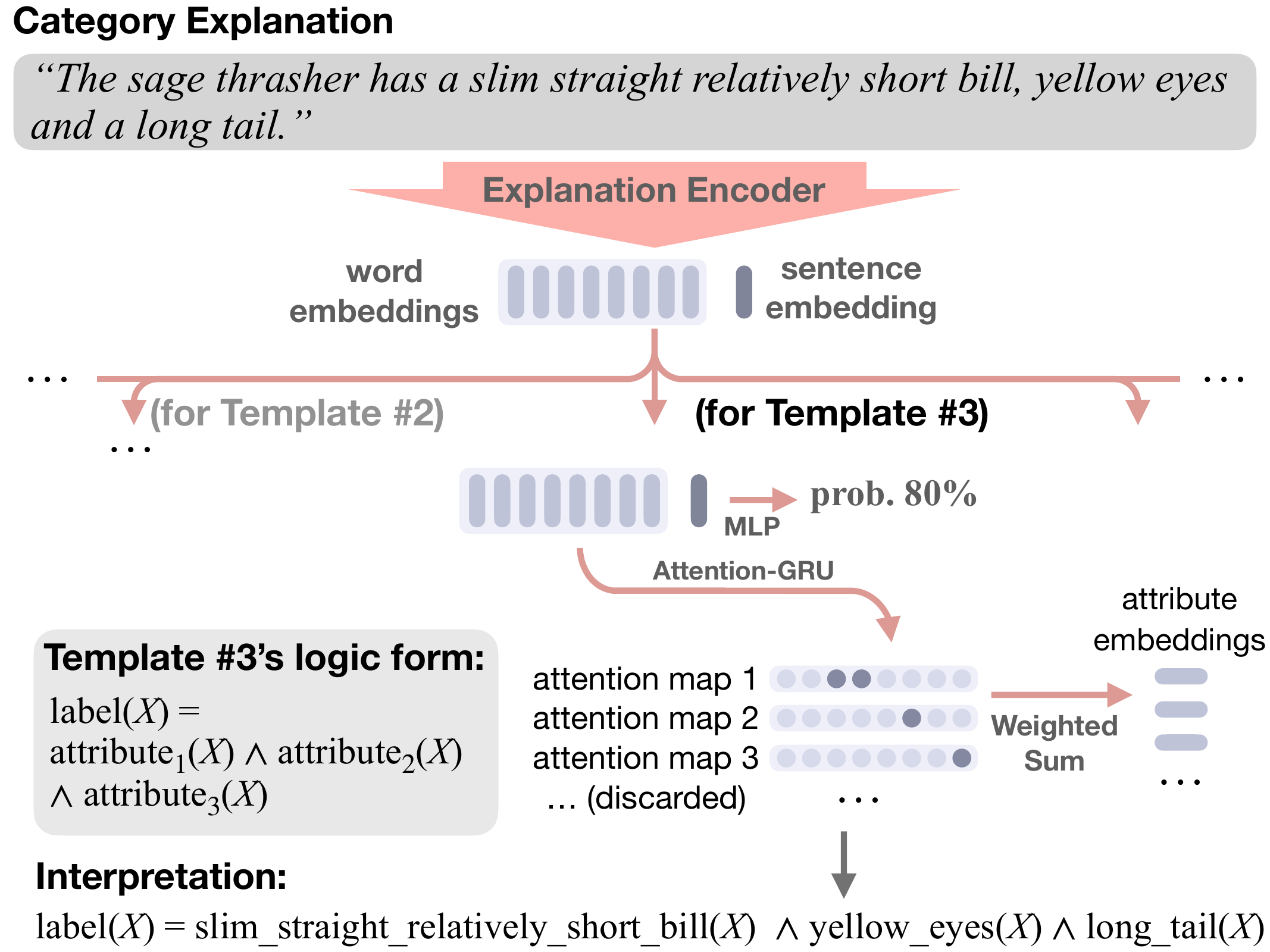}

\caption{We parse each explanation into its logical structure. For each template, we predict its probability and attribute embeddings given by attention-based weighted sum.
}

\vspace{-4mm}
\label{fig:parsing}
\end{figure}
\section{Logical Parsing and Reasoning}

% \heng{\sout{I added the section title back, and your current section 4 logical parsing and reasoning is empty}}
\label{sec:approach}

Explanation-based classification 
% \heng{\sout{maybe call it as 'explanations based classification'}}
is, in essence, a bilateral matching problem between inputs and explanations. Instead of simply using similarity or entailment scores, in this work we aim at better utilizing the logical structure of natural language explanations. A detailed illustration of our proposed model, \model{}, is shown in Figure~\ref{fig:approach}. At the core of the approach is a 2-stage logical matching process: logical parsing of the explanation (Figure~\ref{fig:approach}(d)) and logical reasoning on explanation and inputs to obtain the classification scores (Figure~\ref{fig:approach}(e)).
Rather than using sentence embeddings, our approach focuses more on the logical structure of language explanations, setting it apart from logic-agnostic baselines such as ExEnt and RoBERTa-sim (which is based on sentence embedding similarity). To the best of our knowledge, ours is the first attempt to utilize logical structure in zero-shot classification benchmarks, and it also serves as a proof of concept for the importance of language compositionality.
In the following part of this section we will describe these two stages. More implementation details including input representation can be found at Section~\ref{sec:clues} and \ref{sec:other_modalities}.

% \subsection{Input Representation}
% \paragraph{Modality-Specific Input Encoder} $E_\Phi$ is a learnable module which converts inputs $I_i$ into a uniform representation $X=E_\Phi(I_i)=(x_1, x_2, \cdots, x_K)$. On structured and textual data inputs, $E_\Phi$ uses pre-trained language models such as BERT~\citep{kenton2019bert} as model backbone. The structured data needs to be first re-written as a textual sequence before being encoded, which will be detailed in the Experiments section. On image data, we use pre-trained image networks such as ResNet-101~\citep{he2016deep} and CLIP~\citep{radford2021learning} for encoding. In order to obtain a sequence representation $X$, we abandon the final linear and pooler layers of ResNet-101 and CLIP, and flatten the last hidden outputs into a sequence.

\subsection{Logical Parsing}

This stage is responsible for detecting attributes mentioned in an explanation as well as recovering the logical structure on top of these attributes. (Figure~\ref{fig:approach}(b) to Figure~\ref{fig:approach}(d)). A more detailed illustration is given in Figure~\ref{fig:parsing}. We divide this parsing into 2 steps:

\paragraph{Step 1: Selecting attribute Candidates}
We deploy a attribute detector to mark a list of attribute candidates in the explanations. Each attribute candidate is associated with an attention map as in Figure~\ref{fig:parsing}. \heng{\sout{show this representation in figure 2?}} First we encode the explanation sentence with a pre-trained language encoder, such as RoBERTa~\citep{liu2019roberta}. This outputs a sentence embedding vector and a sequence of token embedding vectors. Then we apply an attention-based Gated Recurrent Unit (GRU) network ~\citep{qiang2017hierarchical}. Besides the output vector at each recurrent step, attention-based GRU also outputs an attention map over the inputs that is used to produce the output vector. In this work, we use the sentence embedding vector as the initialization vector $h^0$ for GRU, and word embeddings as the inputs. We run GRU for a maximum of $T$ (a hyper-parameter) steps, and get $T$ attention weight maps.
Finally we adopt these attention maps to acquire weighted sums of token features $\{w_t | t\in[1..T]\}$ as attribute embeddings. \heng{\sout{any relation between t and T? it's confusing}}

\paragraph{Step 2: Parsing Logical Structure} The goal of this step is to generate a logical structure over the attribute candidates in the previous step. As shown in Figure~\ref{fig:approach}(d), the logical structure is a binary directed tree with nodes being logical operators \texttt{AND} or \texttt{OR}. Each leaf node corresponds to an attribute candidate. In this work, we need to deal with the problem of undetermined number of attributes, and also allow for differentiable optimization. To this end, we define a fixed list of tree structures within maximum number of $T$ leaf nodes, each resembling the example in Figure~\ref{fig:parsing}. \heng{\sout{need to write clearly what you mean by templates, clarify with examples}} A complete list is shown in Appendix~\ref{appsec:templates}. We compute a distribution on templates by applying an multi-layer perceptron (MLP) with soft-max onto the explanation sentence embedding. This provides a non-negative vector $\boldsymbol{p}$ with sum 1, which we interpret as a distribution over the logical structure templates. If the number of attributes involved in the template is fewer than $T$, we discard the excessive candidates in following logical reasoning steps.
\zhenhailong{\sout{question: how do you fit the candidate attributes into the templates? what if the number of candidate attributes is different with the number of leaf nodes? and how do you decide which candidate attribute go into which leaf node?}}

\subsection{Logical Reasoning}
After getting attribute candidates and a distribution over logical structures, we conduct logical reasoning on the input to get the classification score. An illustration is provided in Figure~\ref{fig:approach}(e).
\paragraph{Step 1: Matching attributes with Inputs}
 We assume that the input is represented as a sequence of feature vectors $X=(x_1,x_2,\cdots,x_K)$. First we define a matching score between attribute embedding $w_t$ and input $X$ as the maximum cosine similarity:
\[
    \text{sim}(X,w_t) \colon= \max_k \cos(x_k,w_t).
\]

\paragraph{Step 2: Probabilisitc Logical Reasoning} This step tackles the novel problem of reasoning over logical structures of explanations. \heng{\sout{is this logical tree based reasoning new? if so emphasize the novelty}} During reasoning, we iterate over each logical tree template  and 
% \textit{assume} that this is the correct logical structure, and 
walk along the tree bottom-up to get the intermediate reasoning scores node by node. First, for leaf nodes in the logical tree (which are associated with attributes), we use the attribute-input matching scores in the previous step as their intermediate scores. Then, for a non-leaf node, if it is associated with an \texttt{AND} operator, we define its intermediate score as $\min(s_1,s_2)$ with $s_1$ and $s_2$ following common practice~\citep{mao2019neuro}. If the non-leaf node is associated with an \texttt{OR} operator instead, we use $\max(s_1,s_2)$ as the intermediate score. The intermediate score of the root node $s_{root}$ serves as  the output reasoning score.
Note that we generated a \textit{distribution over logical structures} rather than a deterministic structure. Therefore, after acquiring the reasoning scores on each structure, we use the probability distribution weight $\boldsymbol{p}$ to sum up the scores $\boldsymbol{s}$ of all structures. The resulting score is then equivalent to probabilistically logical reasoning over a distribution of logical structures.
\[
    s_{expl} = \boldsymbol{p}^\top \boldsymbol{s}
\]

We also consider that some explanations might be more or less certain than others. When using words like ``maybe'', the explanation is less certain than another explanation using word ``always''. We model this effect by associating each explanation with a certainty value $c_{certainty}$, which is produced by another MLP on the explanation sentence embedding. So we scale the score $s_{expl}$ with $c_{certainty}$ in logit scale: 
\[
    s_{scaled} = \sigma(c_{certainty} \cdot \text{logit}(s_{expl}))
\]
Intuitively, the training phase will encourage the model to learn to assign each explanation a certainty value that best fits the classification tasks.
\heng{\sout{unclear how the certainty is computed}}

\paragraph{Step 3: Reasoning over Multiple Explanations} There are usually multiple explanations associated with a category. In this case, we take the maximum $s_{scaled}$ over the set of explanations as the classification score for this category.

\begin{table}[t!]
\centering
\small
\linespread{1}

\resizebox{\linewidth}{!}{
% \begin{tabular}{l|cccc}

% \toprule

% Top-1 acc/\% & \textbf{\model{}} & \textbf{ExEnt} & \textbf{ExEnt-BERT} & \textbf{ExEnt-GPT2} \\
% \midrule
% CLUES-Real & \textbf{57.4} & 54.8 & 46.4 & 43.8 \\
% \midrule
% \hspace{3mm}+pre-training & \textbf{55.2} & 52.7 & 50.5 & 52.4 \\

% Top-1 acc/\% & \textbf{\model{}} & \textbf{\model{}-plain}& \textbf{ExEnt} & \textbf{RoBERTa-sim} \\
% \midrule
% CLUES-Real & \textbf{57.4} & 45.8 & 54.8 & 45.1 \\
% \midrule
% \hspace{3mm}+pre-training & \textbf{55.2} & 49.8 & 52.7 & 46.3 \\

\begin{tabular}{c|ccc}
Top-1 acc/\% 
& CLUES-Real & + pre-training \\
\midrule

% \textbf{RoBERTa-sim} & 54.8 & 52.7 \\
% \textbf{ExEnt} & 45.1 & 46.3 \\

\textbf{ExEnt}  & 54.8 & 52.7 \\
\textbf{RoBERTa-sim} & 45.1 & 46.3 \\

\midrule 

\textbf{\model{}-plain} & 45.8 & 49.8 \\
\textbf{\model{}} & \textbf{57.4} & \textbf{55.2} \\

\bottomrule

\end{tabular}
}

\caption{Cross-task generalization results on CLUES dataset~\citep{menon2022clues}. The first row of results are acquired by only fine-tuning on CLUES-Real, and the second row shows results with additional pre-training on CLUES-Synthetic.
% \chihan{better baselines than GPT2 and BERT to be added, like similarity, ignoring logical structure, entailment?}
}
% \vspace{-4mm}
\label{tab:clues_main_results}
\end{table}

\section{Experiments on Zero-Shot Classification}
In this section we conduct in-depth analysis of our proposed approach towards zero-shot classification with explanations. We start with a latest benchmark, CLUES~\citep{menon2022clues}, which evaluates the performance of classifier learning with natural language explanations. CLUES focuses on the modality of structured data, where input data is a table of features describing an item. This data format is flexible enough for computers on a wide range of applications, and also benefits quantitative analysis in the rest part of this section.
% In the next section (\ref{sec:other_modalities}), we also extend to visual modality.
\label{sec:clues}
\begin{table*}[t!]
\centering
\small
\linespread{1}

\setlength{\tabcolsep}{1mm}{

\NewDocumentCommand{\blueback}{ mO{} }{\colorbox{blue!15}{#1}}
\NewDocumentCommand{\greenback}{ mO{} }{\colorbox{green!30}{#1}}

\begin{tabular}{p{17mm}|p{60mm}|p{78mm}}

\toprule
\textbf{Task} & \textbf{Natural Language Explanation} & \textbf{Interpreted Logical Structure} \\
\midrule

car-evaluation
&
\textit{Cars \blueback{with higher safety} \greenback{and capacity} are highly acceptable for resale.}
&
% ($c_{certainty}$=0.74) 
$~\text{Label}(X)=\text{\blueback{with\_higher\_safety}}(X)\land \text{\greenback{and\_capacity}}(X)$
\\

% \hline

% mushroom
% &
% \textit{\blueback{Foul smelling} are \greenback{Poisonous}. 7 of 7 rows,with no deviation.}
% &
% Label$(X)$ = $\blueback{Foul\_smelling}(X) \land \greenback{Poisonous}(X)$
% \\

\hline

indian-liver-patient
&
\textit{Age \blueback{group above 40} \greenback{ensures liver} patient}
&
Label$(X)$ = $\blueback{group\_above\_40}(X) \land \greenback{ensures\_liver}(X)$
\\

\hline

soccer-league-type
&
\textit{If the \blueback{league is W}-PSL then its type is women's soccer}
&
Label$(X)$ = $\blueback{league\_is\_W}(X)$
\\

\hline

award-nomination-result
&
\textit{If the name of \blueback{association has 'American'} in it then the result was mostly won.}
&
Label$(X)$ = $\blueback{association\_has\_'American'}(X)$
\\

% \hline
% dry-bean
% &
% \textit{Below the 215.00 equivalent diameter leads to the \blueback{Dermos}-an class.}
% &
% Label$(X)$ = $\texttt{\blueback{Dermos}}(X)$
% \\

\bottomrule

\end{tabular}
}

% \vspace{-2mm}
\caption{Examples of interpreted logical structures learned by \model{}. We randomly select 5 tasks from CLUES dataset, and use the alphabetically first explanation for interpretation. In each logical structure, the words corresponding to the detected attributes are colored in the explanation.
}
\label{tab:program_examples}
\end{table*}
\begin{figure*}[t]
\centering
\includegraphics[width=\textwidth]{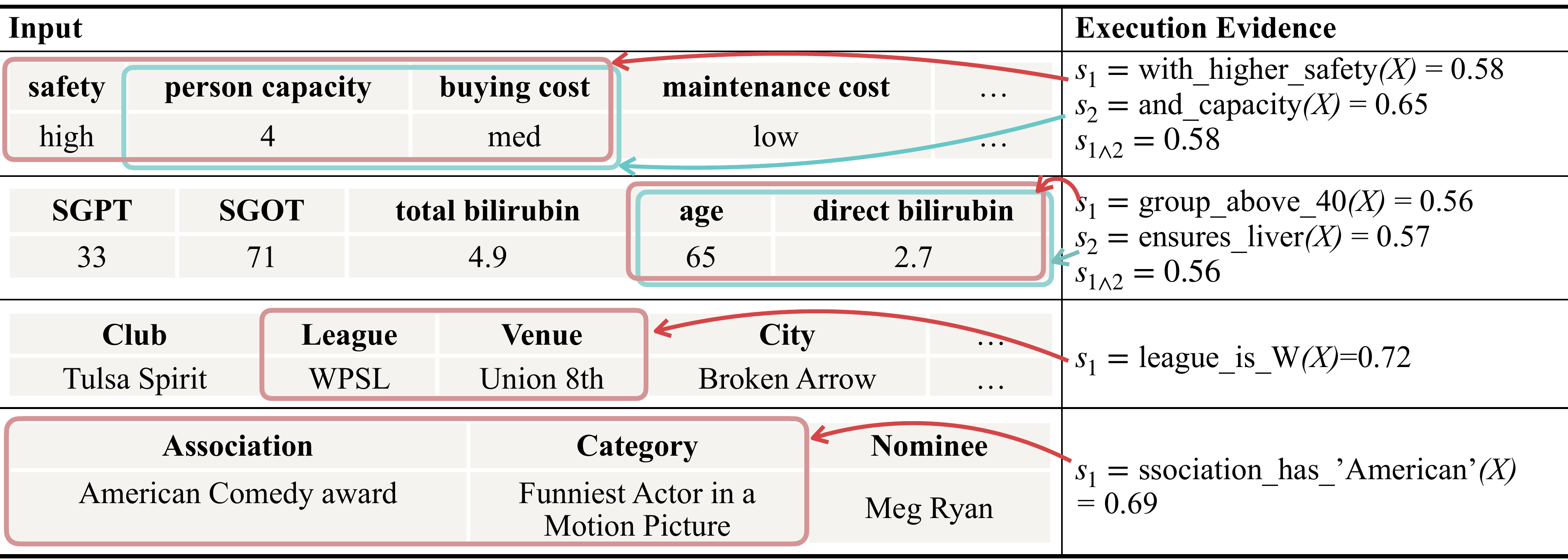}

\vspace{-2mm}
\caption{Examples of logical reasoning evidence. The evidence table cells are linked to attributes with colored arrows.
}

\vspace{-4mm}
\label{fig:execution_evidence}
\end{figure*}

\subsection{CLUES benchmark}
CLUES is designed as a cross-task generalization benchmark on structured data classification. It consists of 36 real-world and 144 synthetic multi-class classification tasks, respectively.
% \heng{what is the data modality for this data set initially?}\hanchi{This dataset is on structured data upon construction. Do you see some problem in adopting this setting?}
The model is given a set of tasks for learning, and then evaluated on a set of unseen tasks.
The inputs in each task constitute a structured table.
Each column represents an attribute type, and each row is one input datum. In each task, for each class, CLUES provides a set of natural language explanations.
% For example, in the ``Mushroom'' task, the columns  are ``odor'', ``stalk-surface-above-ring'', ``gill-color'' and ``ring-type''. The explanations are sentences including \textit{Foul smelling are Poisonous. 7 of 7 rows,with no deviation.}
% We follow the data pre-processing in~\cite{menon2022clues} and convert each input into a text sequence. The text sequence is in the form of ``\texttt{odor | pungent [SEP] ... [SEP] ring-type | pendant}'', where ``\texttt{odor}'' is the attribute type name, and ``\texttt{pungent}'' is the attribute value for this input, and so on.
% Then we encode the sentence with BERT~\citep{kenton2019bert} as inputs $X$.

We follow the data processing in~\citet{menon2022clues} and convert each input into a text sequence. The text sequence is in the form of ``\texttt{odor | pungent [SEP] ... [SEP] ring-type | pendant}'', where ``\texttt{odor}'' is the attribute type name, and ``\texttt{pungent}'' is the attribute value for this input, so on and so forth. For \model{}, we encode the sentence with RoBERTa~\citep{liu2019roberta}\footnote{\url{https://huggingface.co/roberta-base}} and use the word embeddings as input features $X$. More implementation details can be found in Appendix~\ref{appsec:configuration}. We use ExEnt as a baseline, which is an text entailment model introduced in the CLUES paper. ExEnt uses pre-trained RoBERTa as backbone. It works by encoding concatenated explanations and inputs, and then computing an entailment score. We also introduce a similarity-based baseline, RoBERTa-sim, which uses cosine between RoBERTa-encoded inputs and explanations as classification scores. Finally, we compare with \model{}-plain as an ablation study, which ignores the logical structure in \model{} and plainly addes all attribute scores as the overall classification scofre.
\heng{\sout{make it clear you are using ExEnt as baseline}}
\subsection{Zero-Shot Classification Results}
\heng{\sout{really need to put in some qualitative analysis with examples. If you have them in appendix move some of them here since not everyone reads appendix}}
\heng{\sout{I still don't see which results show that your method achieves better interpretability}}

% \heng{\sout{Table 2 is out of margin, fix it}}
% \heng{\sout{What is ExEnt?}}
% \hanchi{\sout{Oh yeah. I just added this part above.}}

Zero-shot classification results are listed in Table~\ref{tab:clues_main_results}. \Model{} outperforms the baseline methods on main evaluation metrics.
To understand the effect of backbound model, we need to note that ExEnt also uses RoBERTa as the backbone model, so the \model{} and baselines do not exhibit a significant difference in basic representation abilities. The inferior performance of RoBERTa-sim compared to ExEnt highlights the complexity of the task, indicating that it demands more advanced reasoning skills than mere sentence similarity.
Furthermore, as an ablation study, \model{} outperforms \model{}-plain, which serves as initial evidence on the importance of logical structure in reasoning.
% Appendix~\ref{appsec:ablation} provides a more detailed ablation study on the effect of logical reasoning complexity.
% Note that due the the gap between CLUES-Real and CLUES-Synthetic, after pre-trained on CLUES-Synthetic some models observe a performance drop. This accords with the observation in the CLUES original paper~\citep{menon2022clues}.
% \hengzhi{\sout{consider to use subsection instead of paragraph}}

\subsection{Effect of Explanation Compositionality}
% \hanchi{\sout{Heng: add a table to show complex/simple examples}}
What causes the difference in performance between \model{} and baselines? To answer this question, we investigate into how the models' performance varies with the compositionality of each task on CLUES.
Table~\ref{tab:compositional_example} provides a pair of examples. An explanations is called ``simple explanation'' if it only describes one attribute, e.g., ``If safety is high, then the car will not be unacceptable.''. Other explanations describe multiple attributes to define a class, e.g., ``Cars with higher safety and medium luggage boot size
are highly acceptable for resale.''. 
We define the latter type as ``compositional explanation''.
% We hypothesize that tasks with higher ratio of compositional explanations require more complex reasoning, 
% \heng{\sout{this point is very good, and needs to be mentioned in the intro}}
% which causes the performance difference between \model{} and ExEnt.
In Figure~\ref{fig:effect_of_complexity} we plot the classification accuracy against the proportion of compositional explanations in each subtask's explanation set.
% \hengzhi{\sout{Could it be a problem that \model{} performance is just stable but not increase as the ratio of compositional explanations increases?}}
% \hanchi{\sout{Good question. How about this explanation?}}
Intuitively, with more compositional explanations, the difficulty of the task increases, so generally we should expect a drop in performance.
Results show that, on tasks with only simple explanations (x-value = 0), both models perform similarly. However, with higher ratio of compositional explanations, \Model{}'s performance generally remains stable, but ExEnt's performance degrades. This validates our hypothesis that \model{}'s performance gain mainly benefits from its better compositional reasoning power.

\begin{figure}[t]
\centering
\includegraphics[width=\columnwidth]{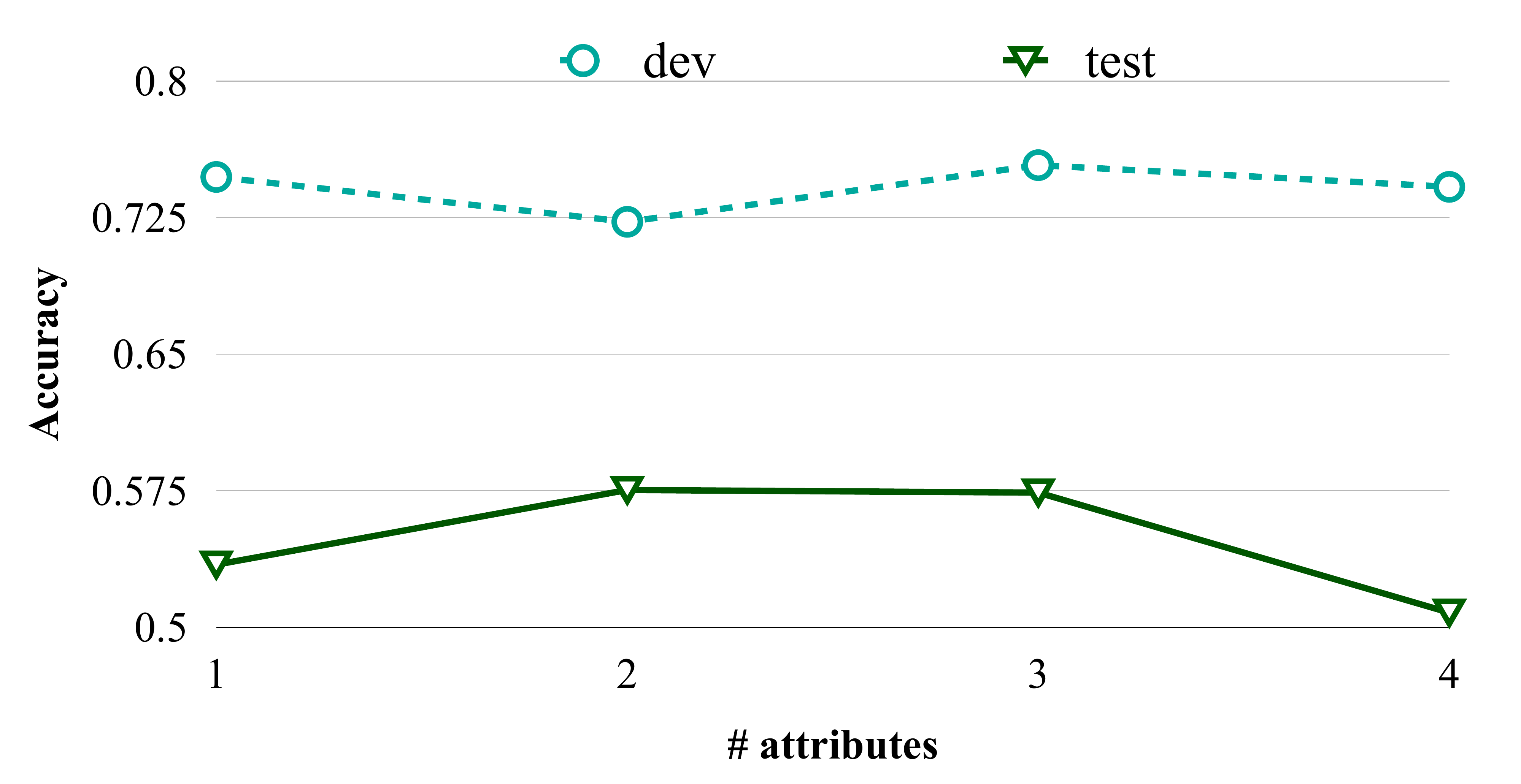}

\caption{The effect of maximum number of attributes $T$ on the classification performance. When $T=1$ the model reduces to a simple similarity-based model.
% The dashed line shows the scores on development set, and solid line shows the test set performance.\xd{the second sentence can be removed since the fig notation provides info.}
}
\label{fig:program_length}
\end{figure}

% \subsection{Effect of Logical Reasoning Complexity}
% \label{appsec:ablation}
To further explore the effect of logical reasoning on model performance. Figure ~\ref{fig:program_length} plots the performance regarding the maximum number of attributes $T$. Generally speaking, when $T$ is larger, \model{} can model more complex logical reasoning process. When $T=1$, the model reduces to a simple similarity-based model without logical reasoning. The figure shows that when $T$ is 2$\sim$3, the model generally achieves the highest performance, which also aligns with our intuition in the section~\ref{sec:approach}. We hypothesize that a maximum logical structure length up to 4 provides insufficient regularization, and \model{} is more likely to overfit the data.
\begin{table}[t]
\centering
\small
\linespread{1}

\begin{tabular}{p{72mm}}

\toprule

\textbf{Compositional Explanation} \\
\midrule
\textit{Cars with \underline{higher safety} and \underline{medium luggage boot size} are highly acceptable for resale.} \\
\midrule

\textbf{Simple Explanation} \\
\midrule
\textit{If \underline{safety is high}, then the car will not be unacceptable.} \\

\bottomrule

\end{tabular}

\vspace{-2mm}
\caption{Examples of a compositional explanation and a simple one in CLUES dataset.}
\label{tab:compositional_example}
\end{table}

% \vspace{-4mm}

\subsection{Interpretability}

% One advantage logical reasoning is that it provides better interpretable intermediate results~\citep{mao2019neuro, yi2018neural}. 
\Model{} is interpretable in two senses: 1) it parses logical structures to explain how the explanations are interpreted, and 2) the logical reasoning   evidence serves as decision making rationales.
To demonstrate the interpretability of \model{}, in Table~\ref{tab:program_examples} and Figure~\ref{fig:execution_evidence} we present examples of the parsed logical structure and reasoning process.
% We compare \model{}'s behavior with ExEnt baseline. 

% More interpretability examples and error analysis are listed in Appendix~\ref{appsec:interpretability}.

The first example of Table~\ref{tab:program_examples} shows that \model{} selects ``\textit{with higher safety}'' and ``\textit{and capacity}'' as attributes candidates, and uses an \texttt{AND} operator over the attributes.
% The middle part of the figure visualizes the logical reasoning evidence.
In Figure~\ref{fig:execution_evidence} correspondingly, two attributes match with columns 1$\sim$3 and 2$\sim$3, respectively.
% the text span \texttt{safety|high[SEP]person capacity|4[SEP]buying cost|med}, and yield scores of 0.65 and 0.58, respectively. Then the \textit{And} operation outputs classification score of 0.58 ($>$ 0.5, indicating a positive prediction).
This example is correctly classified by our model, but mis-classified by the ExEnt baseline.
% This examplifies the efficacy of \model{} on compositional reasoning on explanations.

\begin{figure}
\centering
\includegraphics[width=\columnwidth]{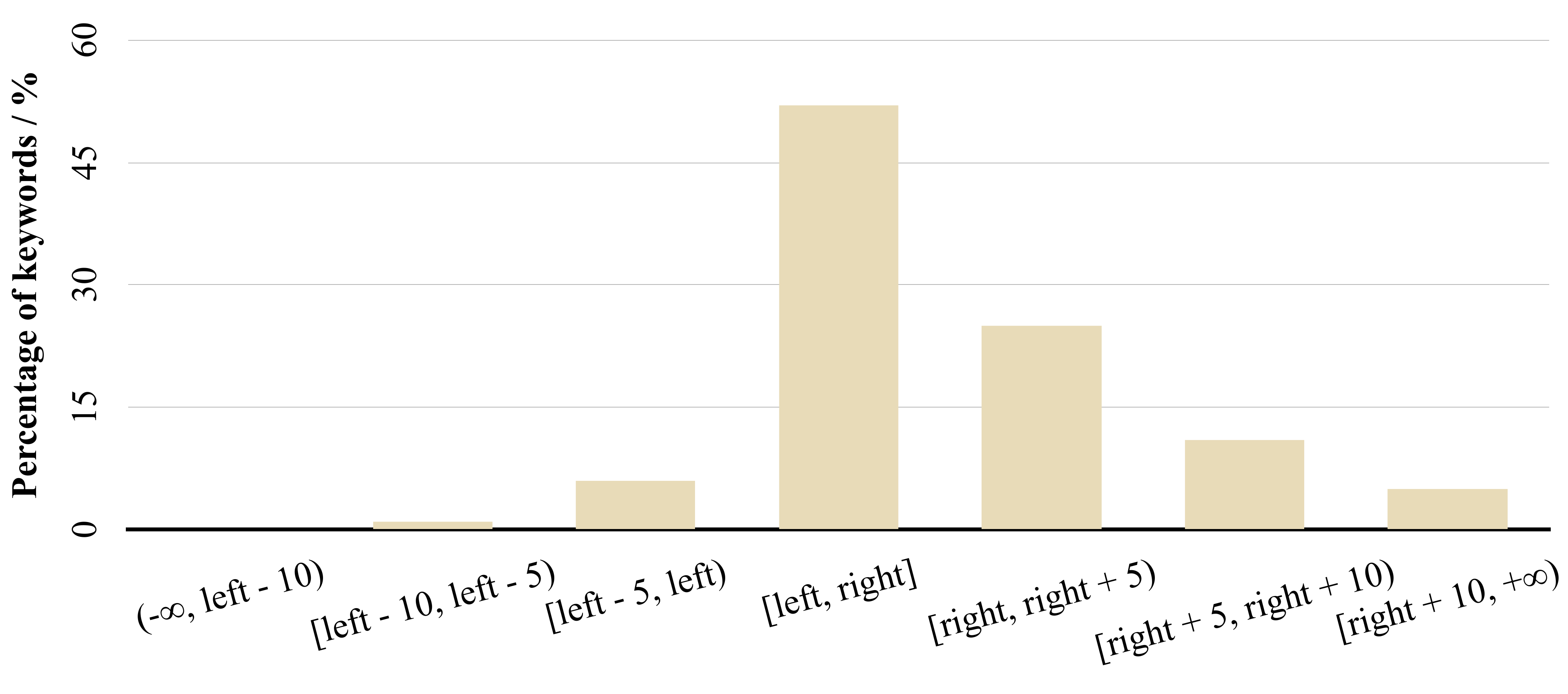}

\caption{The position of detected attributes relative to the expert-annotated keyword spans. Y-axis is the proportion of explanations. Each interval category on x-axis denotes a position range relative to the keyword span in the explanation.
% For example, the first category {\small $[-\infty, left-10]$} stands for the cases where the word with top attention is left to the start-position with a distance greater than 10.
% \xd{\sout{why not merge $[-\infty, left-10]$ and $[left-10,left-5]$ considering you don't discuss $[-\infty, left-10]$ separately}}
}

\vspace{-4mm}
\label{fig:argument_position}
\end{figure}

\begin{figure}
\centering
\includegraphics[width=\columnwidth]{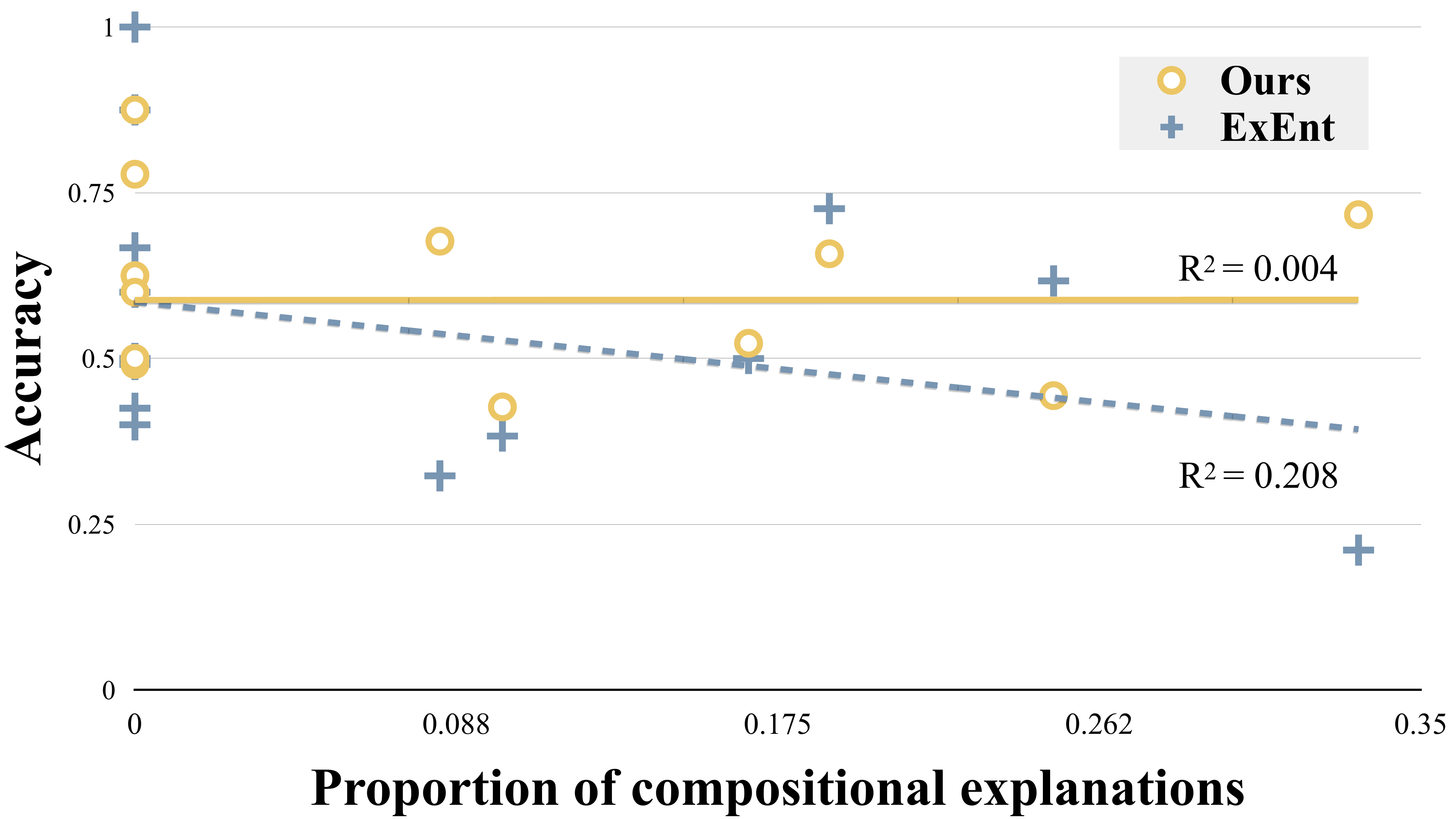}

\caption{The classification accuracy on zero-shot tasks in CLUES plotted against the proportion of compositional explanations. (There are multiple tasks with only simple explanations, so there are multiple points at $x=0$ position.)
% Our model's performance generally remains stable across different types of tasks, but ExEnt performs worse when the task contains more compositional definitions.
}

\vspace{-4mm}
\label{fig:effect_of_complexity}
\end{figure}

To quantitatively evaluate the learned attributes, we manually annotate keyword spans for 100 out of 344 explanations. These spans describe the key attributes for making the explanation. When there are multiple attributes detected, we select the one closest to the keyword span.
% Each keyword span is represented as a pair (start position, end position), which are start and end token positions numbers in the explanation sentences\footnote{by using the \texttt{BERT-base-uncased} tokenizer}.
Then we plot the histogram of the relative position between top-attention tokens and annotated keyword spans in Figure~\ref{fig:argument_position}. From the figure we can see that the majority of top-attention tokens (52\%) fall within the range of annotated keyword spans. 
The ratio increases to 81\% within distance of 5 tokens from the keyword span, and 95\% within distance of 10 tokens.

\begin{figure}
\centering
\includegraphics[width=\columnwidth]{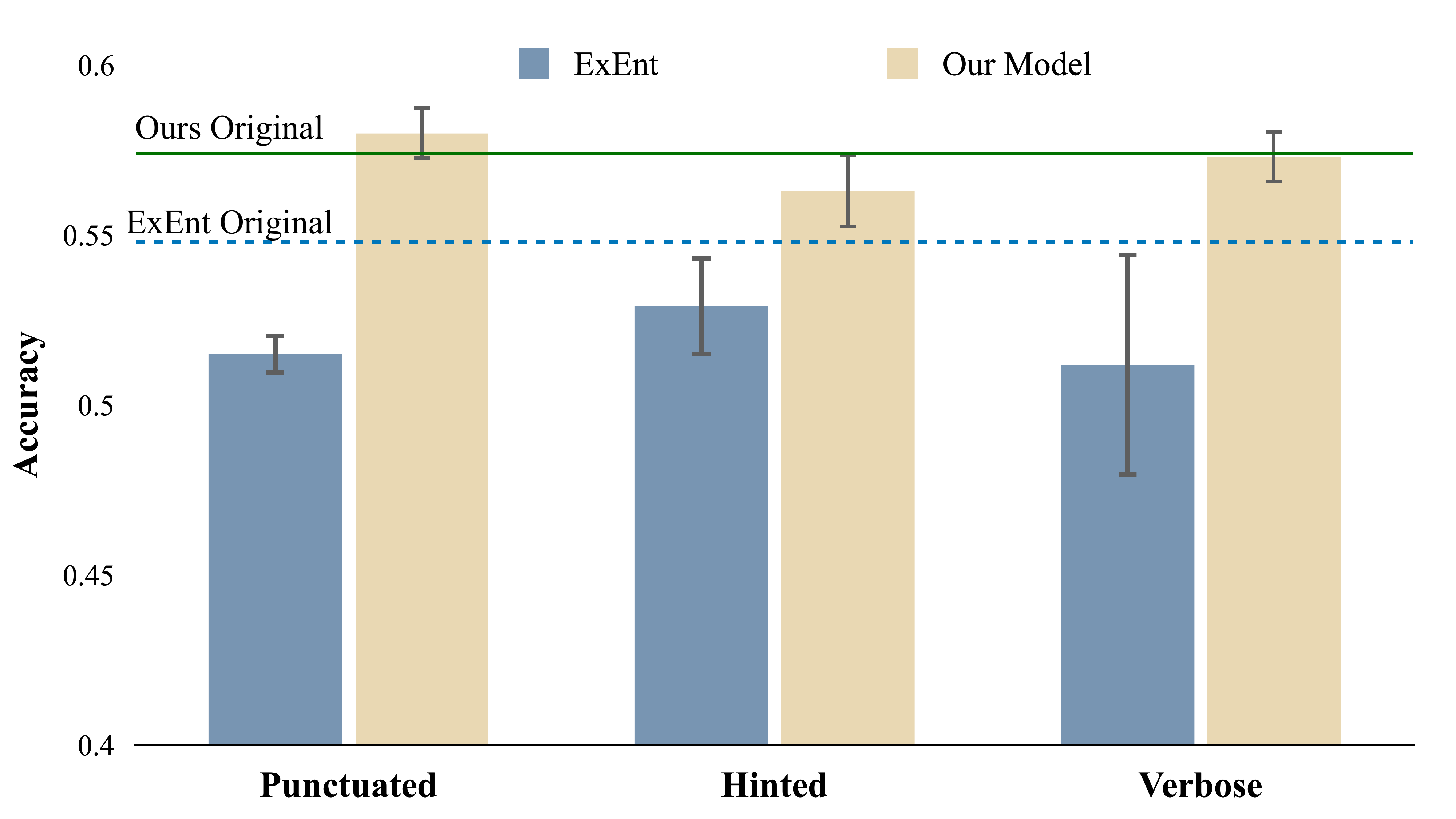}

\caption{The effect of linguistic biases on classifiers. \textit{Punctuated}, \textit{Hinted} and \textit{Verbose} are three types of biasing strategies. The two horizontal lines denote the original performance. Error bars denote standard deviation.}

\vspace{-2mm}
\label{fig:robustness}
\end{figure}

\subsection{Robustness to linguistic bias}

Linguistic biases are prevalent in natural language, which can subtly change the emotions and stances of the text~\citep{field2018framing, ziems2021protect}. Pre-trained language models have also been found to be affected by subtle linguistic perturbations~\citep{kojima2022large} and hints~\citep{patel2021stated}.
% Given that most modern natural language processing models are built on top of these pre-trained language models, they are likely to be susceptible to linguistic biases. 

In this section we investigate how different models are affected by these linguistic biases in inputs. To this end, we experiment on 3 categories of linguistic biases. \textit{Punctuated}: inspired by discussions about linguistic hints in~\cite{patel2021stated}, we append punctuation such as ``?'' and ``...'' to the input in order to change its underlying tone. \textit{Hinted}: we change the joining character from ``\texttt{|}'' to phrases with doubting hints such as ``is claimed to be''. \textit{Verbose}: Transformer-based models are found to attend on a local window of words~\citep{child2019generating}, so we append a long verbose sentence ($\approx$ 30 words) to the input sentence to perturb the attention mechanism. These changes are automatically applied.
% \yi{Are these automatically or manually done?}

Results are presented in Figure~\ref{fig:robustness}. Compared with the original scores without linguistic biases (the horizontal lines), \model{}'s performance is not significantly affected. But ExEnt appears to be susceptible to these biases with a large drop in performance. This result demonstrates that ExEnt also inherits the sensitivity to these linguistic biases from its PLM backbone. By contrast, \model{} is encouraged to explicitly parse explanations into its logical structure and conduct compositional logical reasoning. This provides better inductive bias for classification, and regulates the model from leveraging subtle linguistic patterns.

\subsection{Linguistic Quantifier Understanding} 
\begin{figure}
\centering
\includegraphics[width=1.0\columnwidth]{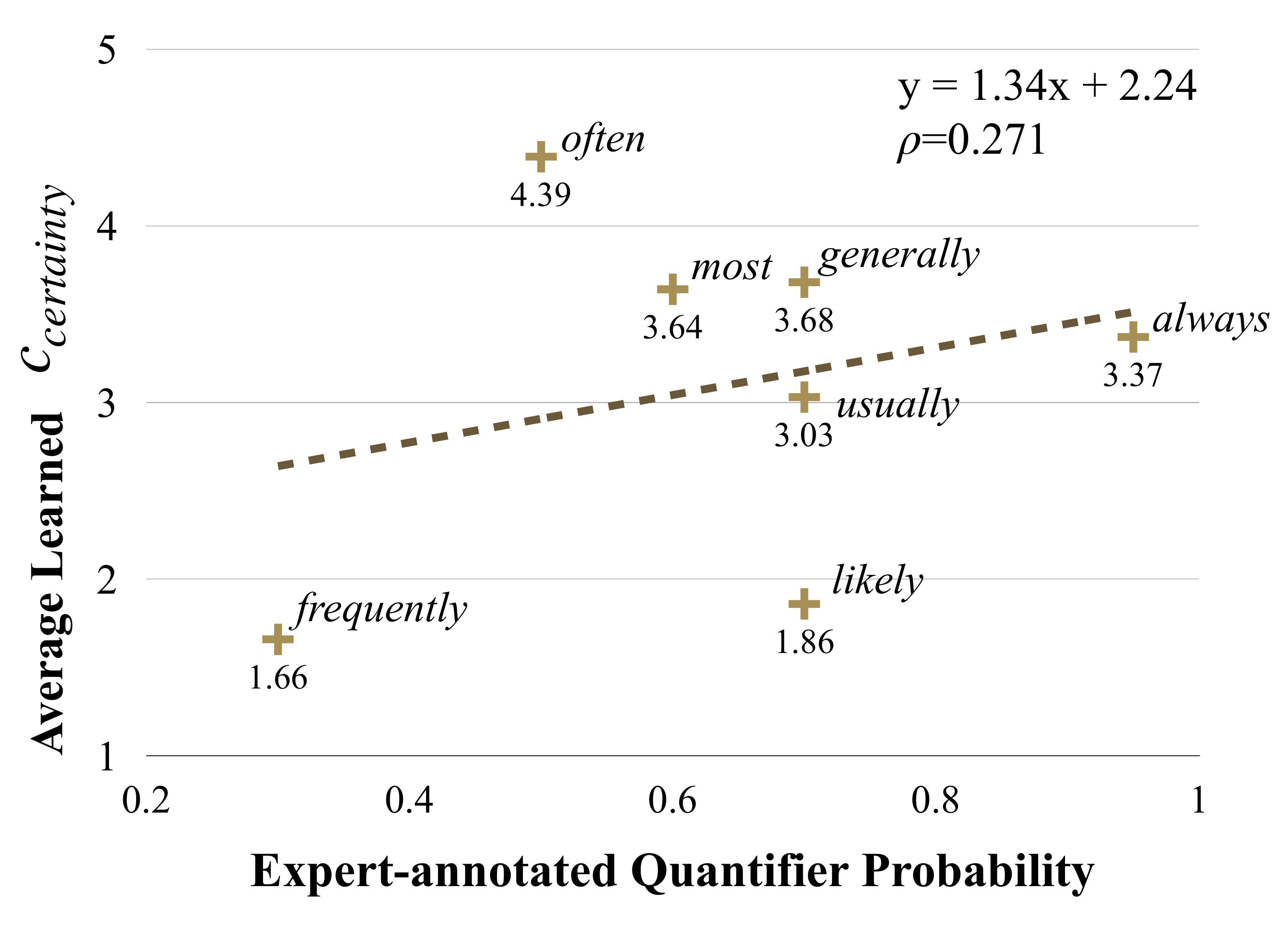}

\caption{Comparison between the learned certainty coefficients $c_{certainty}$ in \model{} and expert annotations in ~\citet{srivastava2018zero}..
% Upper right show the linear regression equation and coefficient of determination ($R^2$).
% \xd{\sout{figure words too small, i cropped the white space around. in the fig, denote what y and $R^2$ is.}}
}

\vspace{-4mm}
\label{fig:quantifiers}
\end{figure}

Linguistic quantifiers is a topic to understand the degree of certainty in natural language~\citep{srivastava2018zero, yildirim2013linguistic}. For example, humans are more certain when saying something \textit{usually} happens, but less certain when using words like \textit{sometimes}.
We observe that the certainty coefficient $c_{certainty}$ that \model{} learns can naturally serve the purpose the of modelling quantifiers.
% \hengzhi{\sout{do we have a explanation for certainty coefficient $c_{certainty}$ before?}}\chihan{\sout{On I forgot. I added this one in Section~\ref{sec:approach}.}}
We first detect the existence of linguistic quantifiers like \textit{often} and \textit{usually} by simply word matching. % listed in Table~\ref{tab:quantifier_extraction}.
Then we take the average of $c_{certainty}$ on the matched explanations. We plot these values against expert-annotated ``quantifier probabilities'' in~\citep{srivastava2018zero}
% , which indicates how ``probable'' the sentence is supposed to be true with a certain quantifier.
in Figure~\ref{fig:quantifiers}. Results show that $c_{certainty}$ correlates positively with ``quantifier probabilities'' with Pearson correlation coefficient value of 0.271. In cases where they disagree, our quantifier coefficients also make some sense, such as assigning \textit{often} a relatively higher value but giving \textit{likely} a lower value.
% \heng{\sout{this paragraph is very hard to follow, and unclear what the point is}}
\hengzhi{and do we need a comparison for this part using the baseline?}
\chihan{I can only think of a most naive way to do this. Let me try it in the next days}

\subsection{Linguistic Quantifier Understanding} Linguistic quantifiers is a topic to understand the degree of certainty in natural language~\citep{srivastava2018zero, yildirim2013linguistic}. For example, humans are more certain when saying something \textit{usually} happens, but less certain when using words like \textit{sometimes}.
We observe that the certainty coefficient $c_{certainty}$ that \model{} learns can naturally serve the purpose the of modelling quantifiers.
% \hengzhi{\sout{do we have a explanation for certainty coefficient $c_{certainty}$ before?}}\chihan{\sout{On I forgot. I added this one in Section~\ref{sec:approach}.}}
We first detect the existence of linguistic quantifiers like \textit{often} and \textit{usually} by simply word matching. % listed in Table~\ref{tab:quantifier_extraction}.
Then we take the average of $c_{certainty}$ on the matched explanations. We plot these values against expert-annotated ``quantifier probabilities'' in~\citep{srivastava2018zero}
% , which indicates how ``probable'' the sentence is supposed to be true with a certain quantifier.
in Figure~\ref{fig:quantifiers}. Results show that $c_{certainty}$ correlates positively with ``quantifier probabilities'' with Pearson correlation coefficient value of 0.271. In cases where they disagree, our quantifier coefficients also make some sense, such as assigning \textit{often} a relatively higher value but giving \textit{likely} a lower value.
% \heng{\sout{this paragraph is very hard to follow, and unclear what the point is}}
\hengzhi{and do we need a comparison for this part using the baseline?}
\chihan{I can only think of a most naive way to do this. Let me try it in the next days}

% \section{Extending to Textual and Visual Inputs}
\section{Extending to Visual Inputs}
\label{sec:other_modalities}
\begin{table}[t!]
\resizebox{\columnwidth}{!}{
\centering
% \small
\linespread{1}

\setlength{\tabcolsep}{1mm}{
\begin{tabular}{c|c|ccc}

\toprule
% \textbf{CUB-Explanations}
& \textbf{Model} & $ACC_U$ & $ACC_S$ & $ACC_\textbf{H}$ \\
\midrule

\multirow{2}{20mm}{w/o VLPMs}
& TF-VAEGAN$_{\mathit{expl}}$ & 4.7 & 39.1 & 8.3 \\
% & Similarity & 4.7 & \textbf{61.2} & 8.8 \\
& \Model{} (ours) & \textbf{6.6} & \textbf{51.1} & \textbf{11.7} \\

\midrule

\multirow{3}{20mm}{w/ VLPMs}
% & CLIP$_{original}$ & 41.5 & 45.9 & 43.6 \\
& CLIP$_{linear}$ & 34.3 & 41.2 & 37.4 \\
& CLIP$_{finetuned}$ & 29.9 & \textbf{66.9} & 41.3 \\
& \Model{}$_{CLIP}$ (ours) & \textbf{39.1} & 65.8 & \textbf{49.1} \\

% \textbf{Model} & \textbf{\Model{}} & \textbf{Similarity} & ZSL\textunderscore TF-VAEGAN & \textbf{\Model{}} & CLIP \\

\bottomrule

\end{tabular}
}
}

\vspace{-2mm}
\caption{Generalized zero-shot classification results (in percentage) on CUB-Explanations dataset.
% $ACC_U$ denotes accuracy on unseen categories, $ACC_S$ denotes accuracy on seen categories, and $ACC_\textbf{H}$ is the harmonic average of the them. 
% The lower and upper parts show models with and without parameters from the Vision Language Pretrained Models (VLPMs).
% TF-VAEGAN$_{def}$ is our re-implementation to adapt to
% \heng{\sout{what do you mean by 'out adaptation'?}}
% CUB-Definitions.
% \Model{}$_{CLIP}$ uses visual and text encoders from CLIP as model backbone.
}
\vspace{-4mm}
\label{tab:cub_main_results}
\end{table}

% Natural language explanations are prevalent in various applications: we use language explanations to define abstract terminologies and describe real-world objects. Taking this observation, in this section we evaluate whether \model{} can be extended to other modalities.

Natural language explanations are prevalent in other applications as well. Taking this observation, in this section we evaluate whether \model{} can be extended to visual domain.
\subsection{Datasets}
\heng{\sout{talk a little bit about the domains used in these datasets}}

Due to lack of datasets on evaluating zero-shot classification with compositional natural language explanations, we augment a standard visual classification datasets with manually collected explanations. Specifically, we select CUB-200-2011~\citep{wah2011caltech}, a bird image classification, as the recognition of birds benefits a lot from their compositional features (such as colors, shapes, etc.).

% Due to lack of datasets on evaluating zero-shot classification with compositional natural language explanations, we augment two standard classification datasets on text and image classification with manually collected explanations. Specifically, we select CUB~\citep{wah2011caltech}, a bird image classification, as the recognition of birds benefits a lot from their compositional features (such as colors, shapes, etc.).
% We also use a legal text dataset, ECtHR~\citep{branting2021scalable}, as experts readily provides official definitions of the legal terms, and explanations are important in legal domain.

\paragraph{CUB-Explanations} We build a CUB-Explanations dataset based on CUB-200-2011, which originally includes $\sim$ 12k images with 200 categories of birds. 150 categories are used for training and other 50 categories are left for zero-shot image classification.
In this work, we focus on the setting of zero-shot classification using natural language explanations.
Natural language explanations of categories are more efficient to collect than the crowd-sourced feature annotations of individual images. They are also similar to human learning process, and would be more challenging for models to utilize.
To this end, we collect natural language explanations of each bird category from Wikipedia. These explanations come from the short description part
% \footnote{As defined in \url{ https://en.wikipedia.org/wiki/Wikipedia:Short_description}}
and the \textit{Description}, \textit{Morphology} or \textit{Identification} sections in the Wikipedia pages.
We mainly focus on the sentences that describe visual attributes that can be recognized in images (e.g. body parts, visual patterns and colors).
Finally we get 1$\sim$8 explanation sentences for each category with a total of 991 explanations.
%CUB-200-2011 originally includes $\sim$ 12k images involving 200 categories of birds, and evaluates models' performance on image classification. In zero-shot classification setting, data of 150 classes are provided for training and other 50 classes are left for evaluation.
% \heng{\sout{so you actually enriched this dataset? add it into contributions}}

% To get input features $X$ for \model{}, we use a pretrained visual encoder (we experiment with both ResNet-101~\citep{he2016deep} and CLIP~\citep{radford2021learning}) to obtain image patch representation vectors. These vectors are then flattened as a sequence and used as visual input $X$.
For evaluation, we adopt the three metrics commonly used for generalized zero-shot learning: $ACC_U$ denotes accuracy on unseen categories, $ACC_S$ denotes accuracy on seen categories, and their harmonic average $ACC_\textbf{H}=\frac{2ACC_UACC_S}{ACC_U + ACC_S}$.

% \paragraph{ECtHR-Explanations} In legal domain the demand for transparent and robust classification is higher ~\citep{branting2021scalable}. This is also a representative domain where humans learn knowledge by reading language definitions.
% So we build a variant dataset ECtHR-Explanations based on ECtHR dataset ~\citep{chalkidis2021paragraph}, which is a recent dataset containing allegations of states violating the European Convention of Human Rights (ECHR). 
% The dataset contains 11k cases of allegations in total. Each case consists of multiple natural language paragraphs describing the facts in the case, and is mapped to a list of allegedly violated ECHR articles. We use the main text under each ECHR article 
% % \heng{\sout{unclear what is language contents, reword it}}
% \footnote{\url{https://www.echr.coe.int/documents/convention_eng.pdf}} as the explanation sentences, thus the name ECtHR-Explanations.

% For evaluation, we adopt a ``leave-one-out'' evaluation method: at each time, one category (article) is left out  for zero-shot evaluation while the remaining categories are used for training.
% % \yi{This part is not too clear to me: does category consists of multiple ECHR violation articles (if so, is this explained somewhere?)}
% Finally we use the averages of F1 scores as evaluation metric.

% Each input is a long document containing multiple paragraphs, so we follow~\cite{chalkidis-etal-2022-lexglue} to use paragraph embeddings as the input features $X$.
%because ROC-AUC score is more stable and does not result in lots of zero scores as F1 metric do.

\subsection{Experiment Setting and Baselines}
% In all experiments, we process the inputs into feature vector sequences $X=(x_1, x_2, \cdots, x_k)$ to provide a uniform representation across modalities.
% On CLUES dataset, this is done by following the data processing in~\citet{menon2022clues} and convert each input into a text sequence. The text sequence is in the form of ``\texttt{odor | pungent [SEP] ... [SEP] ring-type | pendant}'', where ``\texttt{odor}'' is the attribute type name, and ``\texttt{pungent}'' is the attribute value for this input, so on and so forth. Then we encode the sentence with BERT~\citep{kenton2019bert} and use the word embeddings as input features $X$. To set the maximum length of programs, we manually inspect explanations in CLUES dataset, and observe that a maximum length of 5 covers most cases.
On CUB-Explanations dataset, we use a pretrained visual encoder to obtain image patch representation vectors. These vectors are then flattened as a sequence and used as visual input $X$. We use ResNet~\citep{he2016deep} as visual backbone for \model{}.
For baselines, we make comparisons in two groups. The first group of models does not use parameters from pre-trained vision-language models (VLPMs). We adapt TF-VAEGAN~\citep{narayan2020latent}\footnote{\url{https://github.com/akshitac8/tfvaegan}}, a state-of-the-art model on the CUB-200 zero-shot classification task, to use RoBERTa-encoded explanations as auxiliary information. This results in the baseline TF-VAEGAN$_{expl}$. The second group of models are those using pre-trained VLPMs. The main baseline we compare with is CLIP~\citep{radford2021learning}\footnote{\url{https://github.com/openai/CLIP}}, which is a well-performed pretrained VLPM. We build two of its variants: CLIP$_{linear}$, which only fine-tunes the final linear layer and CLIP$_{finetuned}$, which fine-tunes all parameters on the task. For fairer compasion, in this group we also replace the visual encoder with CLIP encoder in our model and get \model{}$_{CLIP}$.

% On the ECtHR-Explanations dataset, for baselines, we use Legal-BERT and hierarchical Legal-BERT as main baselines. They have reported SotA performance on legal-domain datasets \citep{chalkidis-etal-2022-lexglue}. We use two Legal-BERT encoders to encode the explanations and inputs individually, and compute the dot-product between them. The maximal dot-product across explanation sentences is used as classification score for each class. As classification based on explanations shares similarity with natural language inference (NLI), we also add two NLI baselines. BART-large-MNLI~\citep{lewis2020bart} is a powerful model pretrained on NLI tasks, and we fine-tune it on ECtHR-Explanations. T0~\citep{sanh2021multitask} is a strong large pretrained language model capable of a series of tasks, and is used as a zero-shot baseline here without further training due to its large size.
% For \model{}, we follow~\citet{chalkidis-etal-2022-lexglue} and adopt a hierarchical model structure. First the pre-trained language encoder encodes each paragraph as one vector. Then these vectors are concatenated into a vector sequence and used as $X$. To adapt our model to this setting, we also replace the text encoder with Legal-BERT and get \model$_{Legal-BERT}$.
\subsection{Classification Results}

% \paragraph{Results on CUB-Explanations}
Results are listed in Table~\ref{tab:cub_main_results}
% and Table~\ref{tab:ecthr_main_results}
. On CUB-Explanations \model{} achieves the highest $ACC_U$ and $ACC_\textbf{H}$ both with and without pre-trained vision-language parameters.
Note that fine-tuning all parameters of CLIP makes it fit marginally better on seen classes, but sacrifices its generalization ability. Fine-tuning only the final linear layer (CLIP$_{linear}$) provides slightly better generalizability on unseen categories, but it is still lower than our approach.

% On ECtHR-Explanations dataset, our model outperforms the SotA baseline Hierarchical Legal-BERT on all metrics, with gains around 1$\sim$5 percentage points.

\section{Conclusions and Future Work}

\label{sec:conclusions}

In this work, we propose a multi-modal zero-shot classification framework by logical parsing and reasoning on natural language explanations. Our method consistently outperforms baselines across modalities.
% on CUB-200 (image classification), CLUES (structured data classification) and Lex-GLUE and Natural Instructions (text classification).
We also demonstrate that, besides being interpretable, \model{} also benefits more from tasks that require more compositional reasoning, and is more robust against linguistic biases.

There are several future directions to be explored. The most intriguing one is how to utilize pre-trained generative language models for explicit logical reasoning
% , as pre-trained language models have been shown capable of planning~\citep{DBLP:conf/icml/HuangAPM22}
. 
Another direction is to incorporate semantic reasoning ability in our approach, such as reasoning on entity relations or event roles.

\section*{Limitations}
The proposed approach focuses more on logical reasoning on explanations for zero-shot classification. The semantic structures in explanations, such as inter-entity relations and event argument relations, are less touched (although the pre-trained language encoders such as BERT provides semantic matching ability to some extent). Within the range of logical reasoning, our focus are more on first-order logic, while leaving the discussion about higher-order logic for future work.
\section*{Ethics Statement}
This work is related to and partially inspired by the real-world task of legal text classification. As legal matters can affect the life of real people, and we are yet to fully understand the behaviors of deep-learning-based models, relying more on human expert opinions is still a more prudent choice. While the proposed approach can be utilized for automating the process of legal text, care must be taken before using or referring to the result produced by any machine in legal domain.
\subsection*{Acknowledgements}
We would like to thank anonymous reviewers for valuable comments and suggestions.
This work was supported in part by US DARPA KAIROS Program No. FA8750-19-2-1004.
The views and conclusions contained in this document are those of the authors and should not be interpreted as representing the official policies, either expressed or implied, of the U.S. Government. The U.S. Government is authorized to reproduce and distribute reprints for Government purposes notwithstanding any copyright notation here on.

% \section*{Limitations}
% ACL 2023 requires all submissions to have a section titled ``Limitations'', for discussing the limitations of the paper as a complement to the discussion of strengths in the main text. This section should occur after the conclusion, but before the references. It will not count towards the page limit.
% The discussion of limitations is mandatory. Papers without a limitation section will be desk-rejected without review.

% While we are open to different types of limitations, just mentioning that a set of results have been shown for English only probably does not reflect what we expect. 
% Mentioning that the method works mostly for languages with limited morphology, like English, is a much better alternative.
% In addition, limitations such as low scalability to long text, the requirement of large GPU resources, or other things that inspire crucial further investigation are welcome.

% \section*{Ethics Statement}
% Scientific work published at ACL 2023 must comply with the ACL Ethics Policy.\footnote{\url{https://www.aclweb.org/portal/content/acl-code-ethics}} We encourage all authors to include an explicit ethics statement on the broader impact of the work, or other ethical considerations after the conclusion but before the references. The ethics statement will not count toward the page limit (8 pages for long, 4 pages for short papers).

\bibliography{anthology,custom}
\bibliographystyle{acl_natbib}

\appendix

\label{sec:appendix}
\clearpage
\newpage

% \chapter{Appendix}

\section{Appendix}

\subsection{Configuration and Experiment Setting}
\label{appsec:configuration}
We build \model{} on publicly available packages such as HuggingFace Transformers\footnote{\url{https://huggingface.co}, \url{https://github.com/huggingface/transformers}}, where we used model checkpoints as initialization. We train \model{} for 30 epochs in all experiments. In the image classification task on CUB-Explanations, we adopt a two-phase training paradigm: in the first phase we fix both visual encoders and Explanation encoders in $E_\Phi$, and in the second phase we finetune all parameters in \model{}.

Across experiments in this work we use the AdamW~\citep{loshchilov2017decoupled} optimizer widely adopted for optimizing NLP tasks. For hyper-parameters in most experiments we follow the common practice of learning rate$=3e-5$, $\beta_1=0.9, \beta_2=0.999$, $\epsilon=1e-8$ and weight decay$=0.01$. An exception is the first phase in image classification where, as we fix the input encoder, the learnable parameters become much less. Therefore we use the default learning rate$=1e-3$ in AdamW. For randomness control, we use random seed of 1 across all experiments.

In Figure~\ref{fig:effect_of_complexity}, there are multiple data points at $x$-value of 0. Therefore, the data variance on data at $x=0$ is intrinsic in data, and is unsolvable theoretical for \textit{any} function fitting the data series. This causes the problem when calculating $R^2$ value, as $R^2$ measures the extent to which the data variance are ``explained'' by the fitting function. So $R^2$ can be upper bounded by:
$
    R^2 \leq 1 - \frac{Var_{intrinsic}}{Var_{total}}
$. To deal with this problem when measuring $R^2$ metric, we removed the intrinsic variance in data point set $D$ by replacing data points $(0, y_i)\sim D$ with $(0, \frac{1}{n}\sum_{(0,y_i)\sim D}y_i)$ in both series in Figure~\ref{fig:effect_of_complexity} before calculating $R^2$ value.

\subsection{Logical Structure Templates}
\label{appsec:templates}
\begin{table}[t]
\centering
\small
\linespread{1}

\begin{tabular}{l}

\toprule

label$(X)$ = attribute$_1(X)$ \\
\midrule
label$(X)$ = attribute$_1(X) \land$ attribute$_2(X)$ \\
\midrule
label$(X)$ = attribute$_1(X) \lor$ attribute$_2(X)$ \\
\midrule
label$(X)$ = attribute$_1(X) \land$ attribute$_2(X) \land$ attribute$_3(X)$ \\
\midrule
label$(X)$ = attribute$_1(X) \lor$ attribute$_2(X) \lor$ attribute$_3(X)$ \\
\midrule
label$(X)$ = (attribute$_1(X) \land$ attribute$_2(X)) \lor$ attribute$_3(X)$ \\
\midrule
label$(X)$ = (attribute$_1(X) \lor$ attribute$_2(X)) \land$ attribute$_3(X)$ \\

\bottomrule

\end{tabular}

\caption{The list of logical structure templates at maximum attribute number $T=3$.}
\label{tab:all_program_templates}
\end{table}
As the number of valid logical structure templates grows exponentially with maximal attribute numbers $T$, we limit $T$ to a small value, typically 3. We list the logical structure templates in Table~\ref{tab:all_program_templates}.

% \subsection{Effect of Logical Reasoning Complexity}
% \label{appsec:ablation}
% We also explores the effect of logical reasoning on model performance. Figure ~\ref{fig:program_length} plots the performance regarding the maximum number of attributes $T$. Generally speaking, when $T$ is larger, \model{} can model more complex logical reasoning process. When $T=1$, the model reduces to a simple similarity-based model without logical reasoning. The figure shows that when $T$ is 2$\sim$3, the model generally achieves the highest performance, which also aligns with our intuition in the section~\ref{sec:approach}. We hypothesize that a maximum logical structure length up to 4 provides insufficient regularization, and \model{} is more likely to overfit the data.

% \input{figures/program_length}

\subsection{Resources}

We use one Tesla V100 GPU with 16GB memory to carry out all the experiments. The training time is 1 hour for tabular data classification on CLUES, 2 hours for image classification on CUB-Explanations.

\end{document}